\documentclass[10pt, conference, compsocconf]{IEEEtran}
\usepackage{xcolor}

\usepackage{cite}
\ifCLASSINFOpdf
  \usepackage[pdftex]{graphicx} 
\else
\fi
\usepackage[cmex10]{amsmath}
\begin{document}
\title{SealD-NeRF: Interactive Pixel-Level Editing for Dynamic Scenes by Neural Radiance Fields}


\author{\IEEEauthorblockN{Zhentao Huang, Yukun Shi, Neil Bruce, Minglun Gong}
\IEEEauthorblockA{School of Computer Science\\
University of Guelph\\
Guelph, Canada\\
(zhentao, yshi21, brucen, minglun)@uoguelph.ca}
% \and
% \IEEEauthorblockN{Authors Name/s per 2nd Affiliation (Author)}
% \IEEEauthorblockA{line 1 (of Affiliation): dept. name of organization\\
% line 2: name of organization, acronyms acceptable\\
% line 3: City, Country\\
% line 4: Email: name@xyz.com}
}


%Telephone: (800) 555--1212, Fax: (888) 555--1212}


\maketitle

\begin{abstract}
The widespread adoption of implicit neural representations, especially Neural Radiance Fields (NeRF) as detailed by \cite{mildenhall2021nerf}, highlights a growing need for editing capabilities in implicit 3D models, essential for tasks like scene post-processing and 3D content creation. Despite previous efforts in NeRF editing, challenges remain due to limitations in editing flexibility and quality. The key issue is developing a neural representation that supports local edits for real-time updates. Current NeRF editing methods, offering pixel-level adjustments or detailed geometry and color modifications, are mostly limited to static scenes. This paper introduces SealD-NeRF, an extension of Seal-3D for pixel-level editing in dynamic settings, specifically targeting the D-NeRF network \cite{pumarola2021d}. It allows for consistent edits across sequences by mapping editing actions to a specific timeframe, freezing the deformation network responsible for dynamic scene representation, and using a teacher-student approach to integrate changes.

\end{abstract}


% \fi
\IEEEpeerreviewmaketitle

\section{Introduction}
Combining machine learning and geometric insights, neural rendering techniques have emerged as a highly promising approach for generating fresh perspectives of a scene from a limited set of images. Within this group of techniques, one particular standout is Neural Radiance Fields (NeRF) \cite{mildenhall2021nerf}, a method that employs a deep neural network to transform 5D input coordinates, representing spatial location and viewing direction, into both volume density and view-dependent emitted radiance. NeRF and its related methods, such as those documented by \cite{barron2021mip, zhang2020nerf++, fridovich2022plenoxels, muller2022instant}, have demonstrated considerable potential across a range of 3D applications due to their remarkable reconstruction accuracy, high-quality rendering, and efficient memory usage. These technologies are increasingly valuable in fields like 3D reconstruction, generating new viewpoints, and enhancing Virtual and Augmented Reality experiences.

Originally tailored for static settings, NeRF has been extended by D-NeRF \cite{pumarola2021d} to adeptly handle dynamic scenes. This is achieved by integrating a deformation network module with the original neural radiance field framework. To be precise, it can be effectively applied to dynamic scenes that consist of both stationary and moving or deforming objects. It is the first to generate a neural implicit representation for non-rigid and time-varying scenes, trained solely on monocular data without the need for 3D ground-truth supervision or a multi-view camera setting. Other research has also introduced dynamic generalization of NeRF \cite{cao2023hexplane, liu2022devrf, xian2021space, park2021nerfies}.

% \begin{figure*}[]
%   \centering
%   \includegraphics[width=2\columnwidth]{imgs/dnerf.png}
%   \caption{Illustration of the standard D-NeRF streamline \cite{pumarola2021d}.}
%   \label{dnerf}
% \end{figure*}

\begin{figure*}[]
  \centering
  \includegraphics[width=2\columnwidth]{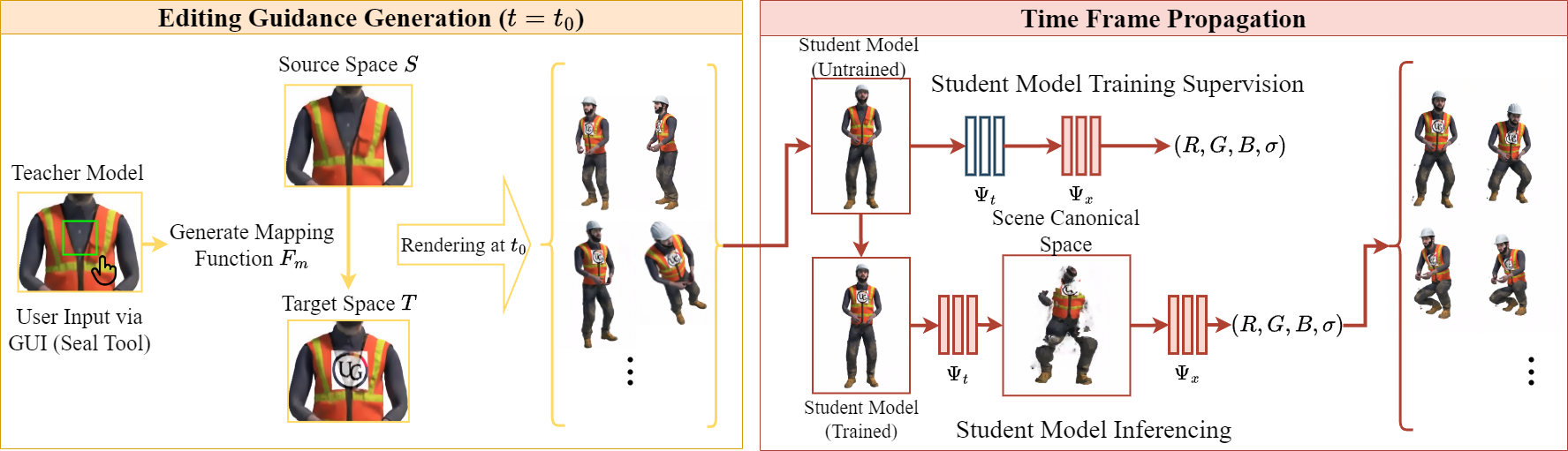}
  \caption{SealD-NeRF contains two main stages: editing guidance generation and time frame propagation. Firstly, the mapping function $F_m$ is generated based on input from the user interface. It is used for mapping the original source space \textit{S} to the target space \textit{T}. The target space is further used for student model training supervision by rendering multiple views. During the student training process, the deformation network $\Psi_t$ is frozen to maintain the object movement. Only the canonical network $\Psi_x$ is optimized to propagate the edit to all the time frames. }
  \label{fig:overview}
\end{figure*}

As implicit representations and the adoption of implicit 3D models gain popularity, there is a growing need for user-friendly editing tools that allow people to interact with these 3D models. Existing research has explored various perspectives including object segmentation \cite{liu2021editing, yang2021learning}, color editing \cite{kuang2023palettenerf}, object removal \cite{liu2022nerf}, etc. These approaches are all focused on object-level editing of NeRF and cannot achieve pixel-level editing. Recently, Seal-3D \cite{wang2023seal} filled this gap by providing an interactive framework for pixel-level editing for NeRF supporting instant preview. It first generates the editing guidance as the teacher model via pre-defined proxy functions. Then, a two-stage student training process including local pretraining and global fine-tuning is adopted to generate the result. However, all the above-mentioned methods only tackle the problems in static scenes. There is no such application on NeRF for dynamic scenes such as D-NeRF \cite{pumarola2021d}. The main challenge is to keep the modification consistent throughout the whole time stream. Manually editing at pixel level from frame to frame is impractical.

This paper introduces a method that enables users to edit an arbitrary frame, with the changes automatically applied to all other frames. The primary goal of this work is to develop an interactive framework for pixel-level editing within the D-NeRF context. This is accomplished by first transferring the user input for editing into the original NeRF space via a predefined proxy function and setting it as the teacher model. Note that there is only one time frame required for this step. There are two sub-networks in the standard D-NeRF framework: a deformation network mapping all scene deformations to a common canonical configuration; and a canonical network regressing volume density and view-dependent RGB color from every camera ray which is identical to a standard NeRF network for static scenes. The key idea of the proposed approach is to keep the original object's movement of the scene by freezing the weights in the deformation network while learning from the teacher model. At the same time, the modification of the scene is adopted purely based on the learning of the canonical network. At present, users have the capability to perform brush edits or seal images onto the surface of objects. Exploring whether this approach can be expanded to include additional actions, such as moving or adding objects, remains an area for future investigation.

The key contributions of this work are outlined below:
\begin{itemize}
    \item Implementation of the D-NeRF network utilizing the Torch-NGP framework.
    \item Development of brush and sealing editing tools for a singular time frame within the D-NeRF model.
    \item Introduction of a technique that permits users to edit a single time frame, ensuring consistency across the entire timeline.
\end{itemize}

\section{Background}

\subsection{Neural Radiance Fields for Dynamic Scenes}
With the popularity of Neural Radiance Fields and the great performance on static scenes, more research has focused on the dynamic scenes in recent years. Weng et al. \cite{weng2022humannerf} proposed a method that transforms a monocular dance video into a free-viewpoint rendering by creating a canonical subject appearance volume and a motion field, allowing synthesis of output views based on the source frame's pose during testing. TiNeuVox \cite{fang2022fast} introduces a radiance field framework with time-aware voxel features, a small deformation network for motion, and a multi-distance interpolation method, demonstrating accelerated dynamic radiance field optimization and high rendering quality in empirical evaluations. Gao et al. \cite{gao2021dynamic} employs scene flow-based regularization, emphasizing its core contribution of enforcing temporal consistency and addressing ambiguity in modeling dynamic scenes with a single observation. Li et al. \cite{li2021neural} introduces Neural Scene Flow Fields, a representation optimized through neural networks to model dynamic scenes. Nerfies \cite{park2021nerfies} and D-NeRF are very similar in the framework structure. However, Nerfies reported better results on real-world scenes. Future work might involve discovering a new network backbone for the proposed method.

\subsection{Neural Radiance Fields Editing}
The task of scene editing has been extensively explored in both computer vision and graphics research. Early methods primarily concentrated on editing a single static view, involving operations such as insertion \cite{zeng2023efficient, kuang2023palettenerf}, relighting \cite{li2020inverse}, and composition \cite{park2021hypernerf}. With the growing demands of scene editing and popularity of neural rendering, recent works attempt to perform editing at various levels: scene-level, object-level, and pixel-level editing.  

Scene-level editing methods aim to modify the overall appearance of a scene, addressing factors such as lighting \cite{guo2020object} and the global color palette \cite{kuang2023palettenerf}. Intrinsic decomposition techniques, exemplified by other studies like \cite{yu2022monosdf, hasselgren2022shape}, seek to disentangle material and lighting fields, enabling the editing of texture or lighting. However, these methods are constrained to modifying global attributes and do not extend to specific objects within the scene. Object-level editing methods employ various strategies to manipulate implicitly represented objects. For instance, Object-NeRF \cite{wang2021neus} utilizes per-object latent codes to decompose the neural radiance field, enabling object-level operations like movement, removal, or duplication. Other methods, such as Liu et al.'s conditional radiance field model \cite{liu2022nerf}, optimize partially based on editing instructions, allowing semantic-level modifications in color or geometry. While some methods use a deformable mesh as an editing proxy, they are constrained to object-level rigid transformations and may not generalize well to arbitrary editing categories. In contrast, pixel-level editing aims to provide precise guidance at the pixel level, without being restricted by predefined objects. NeuMesh \cite{yang2022neumesh} relies on a mesh scaffold, limiting the editing categories and preventing the creation of out-of-mesh geometry structures. Seal-3D \cite{wang2023seal} does not require proxy geometry structures, offering a more direct and extensive approach.

\section{Methodology}
In this section, the methodology of the whole pipeline is presented. Figure \ref{fig:overview} illustrates an overview of our method. The objective is taking a trained D-NeRF model as input and allows the user conduct editing on the scene via the user interface. The teacher model is generated by duplicating the original scene and apply the edits through mapping function. It is further used for supervising the student model training phase. Once the training of the student model is completed, the edit will propagate to all the time frame with consistency.

\subsection{D-NeRF Implementation}
The standard D-NeRF contains two primary components: a deformation network, which maps all scene deformations to a shared canonical configuration, and a canonical network, which performs regression tasks for volume density and view-dependent RGB color for each camera ray. More specifically, the deformation network is optimized to estimate the deformation field between the scene at a specific time frame and the scene in its canonical configuration. Given a 3D point $x$ at time \textit{t}, the deformation network will estimate the displacement $\Delta x$. The given point will be transformed to the canonical space as $x+\Delta x$ and input to the canonical network. The canonical network performs the same as a standard NeRF model capturing the information of all corresponding points in all images. Through this process, it becomes possible to recover the absent information from a particular viewpoint by referencing the canonical configuration, which serves as an anchor interconnecting all the images together. The canonical network is optimized to to take the deformed 3D coordinates and viewing direction as input, and output the emitted color and volume density.

The implementation was developed based on Torch-NGP \cite{Torch-NGP}, which is the same framework for Seal-3D \cite{wang2023seal}. Although the author of Torch-NGP claimed that D-NeRF \cite{pumarola2021d} is already implemented in the newest release, Torch-NGP's implementation deviated from the original D-NeRF paper. The standard D-NeRF consists of two sub-networks: one deformation network $\Psi_t$ taking 3D coordinates and time frame as input then outputs the deformations; one canonical network $\Psi_x$ regressing volume density and view-dependent RGB color from every camera ray perform similar to a standard NeRF. However, while Torch-NGP's implementation has an identical deformation network, it also inputs the 3D coordinates and time frame into the canonical network. Despite the performance of the modified network, inputting the time frame into the canonical network completely negates the initial design motivation of the proposed SealD-NeRF which needs the time frame and canonical network to be separated. The author of the D-NeRF paper released the code implemented in PyTorch, this runs very slow compared to Torch-NGP which is a PyTorch Implementation on CUDA-based Instant-NGP \cite{muller2022instant}. Therefore, a Torch-NGP-based D-NeRF is implemented for this paper.

There are several key points/alternations that need attention: 1) Torch-NGP introduces a deformation loss to the original loss function, corresponding to the magnitude of 3D deformation. This serves as a form of regularization, initially accelerating the search process but potentially hindering convergence in later stages. The assumption behind this is that the object doesn't have large movement during the whole time stream. This is removed in the implementation presented in this paper. 2) According to the D-NeRF paper, for all experiments, the canonical scene is set to be the scene at time $t=0$. This is not the case in the Torch-NGP implementation which leads to non-convergence when deformation loss is also removed. By setting the canonical scene manually, it gives an initial state for the network to converge and all the time frames become interconnected.

The results section details the assessment of our newly implemented D-NeRF. The evidence strongly supports the accuracy of our implementation, affirming that it does not detract from the quality of scene editing outcomes. 

\begin{figure*}
    \centering
    \includegraphics[width=1.8\columnwidth]{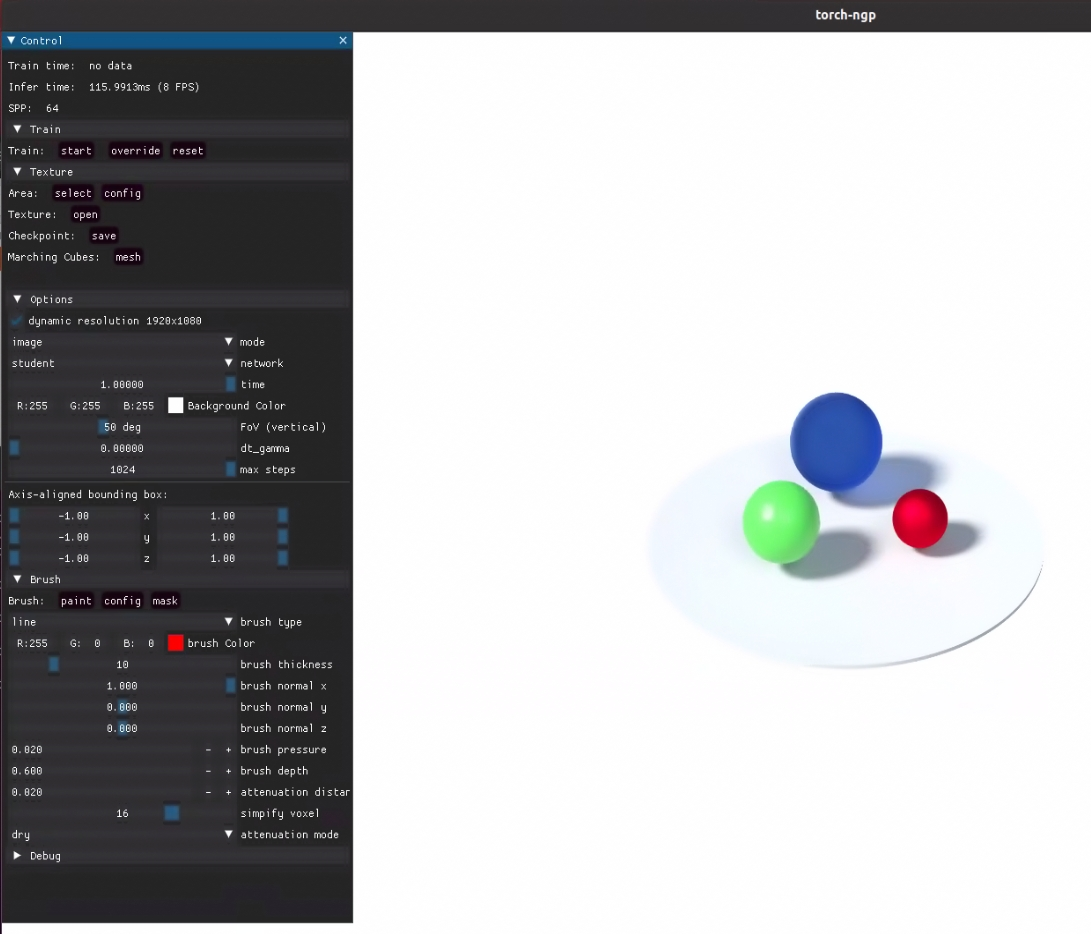}
    \caption{The user interface of the SealD-NeRF, based on Seal-3D \cite{wang2023seal} and Torch-NGP \cite{Torch-NGP}. The user can view the training procedure directly. It allows the user to view any time frame through a control bar. Except for the control parameters of the toolkit, it also allows the user to switch between the student and teacher model for referencing.}
    \label{fig:gui}
\end{figure*}

\subsection{Teacher Model Generation and Time Frame Propagation}
The design of the proposed method is inspired by the process of Seal-3D \cite{wang2023seal}. Given a pretrained D-NeRF network fitting a dynamic scene that serves as a teacher network, an extra NeRF network is initialized with the pretrained weights as a student network. The editing guidance is generated by the teacher network based on the user input instruction on one particular time frame \textit{t}. The student network is optimized by distilling editing knowledge from the editing guidance in that particular time frame. %Maybe add a figure of the streamline

Firstly, pixel-level information is extracted from the interactive editor. The source space \textit{S} is the 3D space for the original NeRF model and the target space \textit{T} is the 3D space for the NeRF model after editing. The target space is warped to the original space by the mapping function $F_m$. The mapping function is used for transforming the target space and its associated directions according to the rules defined by each tool. The desired edited effects $c^\textit{T}, \sigma^\textit{T}$ for each 3D point and view direction at the particular time frame \textit{t} can be acquired by querying the teacher model $f_\theta^\textit{T}$. It's important to emphasize that only the time frame \textit{t} should be employed for the teacher model, as it is the sole frame that incorporates user input instructions. The process can be expressed as:

\begin{equation}
     x^\textit{s}, d^\textit{s} = F^m(x^t, d^t), x^s \in \textit{S}, x^t \in \textit{T}
\end{equation}
   
\begin{equation}
    c^\textit{T}, \sigma^\textit{T} = f_\theta^\textit{T}(x^s, d^s)
\end{equation}

After the teacher model has been set up, the next step is to optimize the student model based on the particular time frame of the teacher model. Within the Seal-3D framework, a local pretraining step is designed for instant preview purposes. However, extending this step to each frame in D-NeRF training is impractical due to the significant computational time it would entail. Therefore, only one global training is applied in our approach. This stage is similar to the standard D-NeRF training, the difference is that the labels are generated by querying the teacher model instead of ground truth image pixels. We use the mean squared error between the pixels rendered by the student model and the teacher model:
\begin{equation}
    \textit{L} = \frac{1}{N}\sum^{N}_{i=1} {\lvert\lvert C^\textit{T} - C^\textit{S} \rvert\rvert}^2_2 
    \label{eq:loss}
\end{equation}

In this equation, $C^\textit{T}, C^\textit{S}$ represents the accumulated pixels from the sampled rays in the teacher and student model, and $N$ is the batch size for each training step. One key point is that this method could propagate errors in the original model (e.g. floating artifacts).

\begin{figure*}
    \centering
    \includegraphics[width=2\columnwidth]{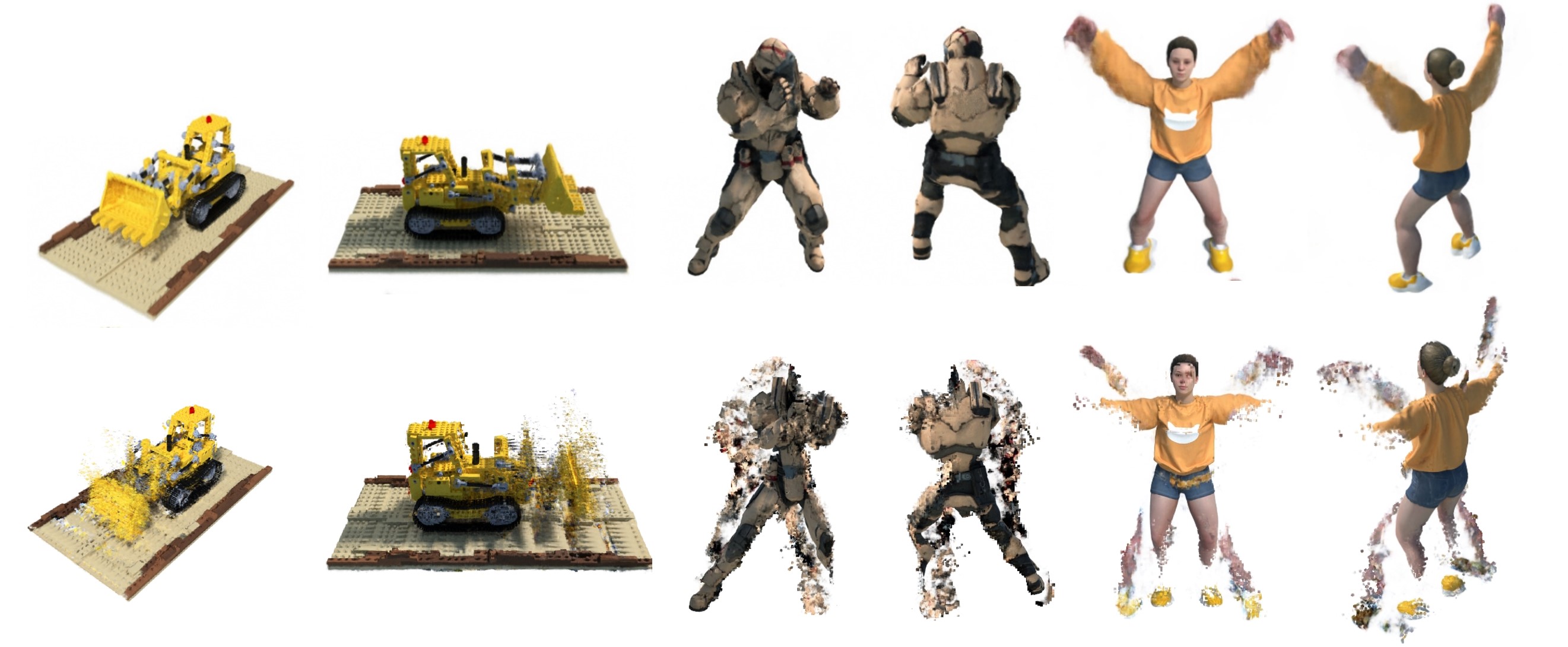}
    \caption{Examples of the canonical spaces of the scenes: "Lego", "Hook", and "Jumping Jacks". \textbf{Top row}: D-NeRF \cite{pumarola2021d} implementation based on PyTorch. \textbf{Bottom row}: our implementation based on Torch-NGP \cite{Torch-NGP}.}
    \label{fig:canonical}
\end{figure*}

To achieve the time frame propagation, the key idea is to keep the current deformations for all the time frames and transform the editing into the canonical space. In this way, the editing will be propagated to the whole time stream since the original D-NeRF already has a converged deformation sub-network $\Psi_t$. In all the experiments, the deformation sub-network is frozen and the canonical network is optimized by inputting rendering output from all angles by the teacher model. In theory, this method is not suited for edits that significantly modify the object's movements. Specifically, any edits necessitating alterations to the existing deformation pattern are likely to be ineffective. This is the main reason why the brush tool and the sealing tool are implemented for now.

\subsection{Brush and Sealing Tools}
The brush and sealing tools mirror those in Seal-3D. The brush tool functions similarly to a sculpt brush in conventional 3D editing, raising or lowering the surface upon which it is used. Users employ the brush to create markings, and the source space \textit{S} is generated through ray casting on the brushed pixels. The brush parameters, including the brush normal \textbf{n} and pressure value \textbf{p} within the range [0, 1], are defined by the user, influencing the resulting mapping:

\begin{equation}
    x^\textit{s} = x^t - \textbf{p}(x^t)\textbf{n}
\end{equation}
\begin{equation}
    F^m:= (x^t, d^t) \Rightarrow (x^s, d^t)
\end{equation}
where $x$ stands for the 3D coordinates.

The sealing tool directly edits color via color space mapping. The spatial mapping is identical and the colors are mapped by the network in HSL space, which helps for preserving shading \cite{wang2023seal}. The user will need to select an area in the space and a 2D image. Then the image will be sealed to the surface of the object and kept consistent throughout the whole time stream.

\begin{figure*}
    \centering
    \includegraphics[width=1.5\columnwidth]{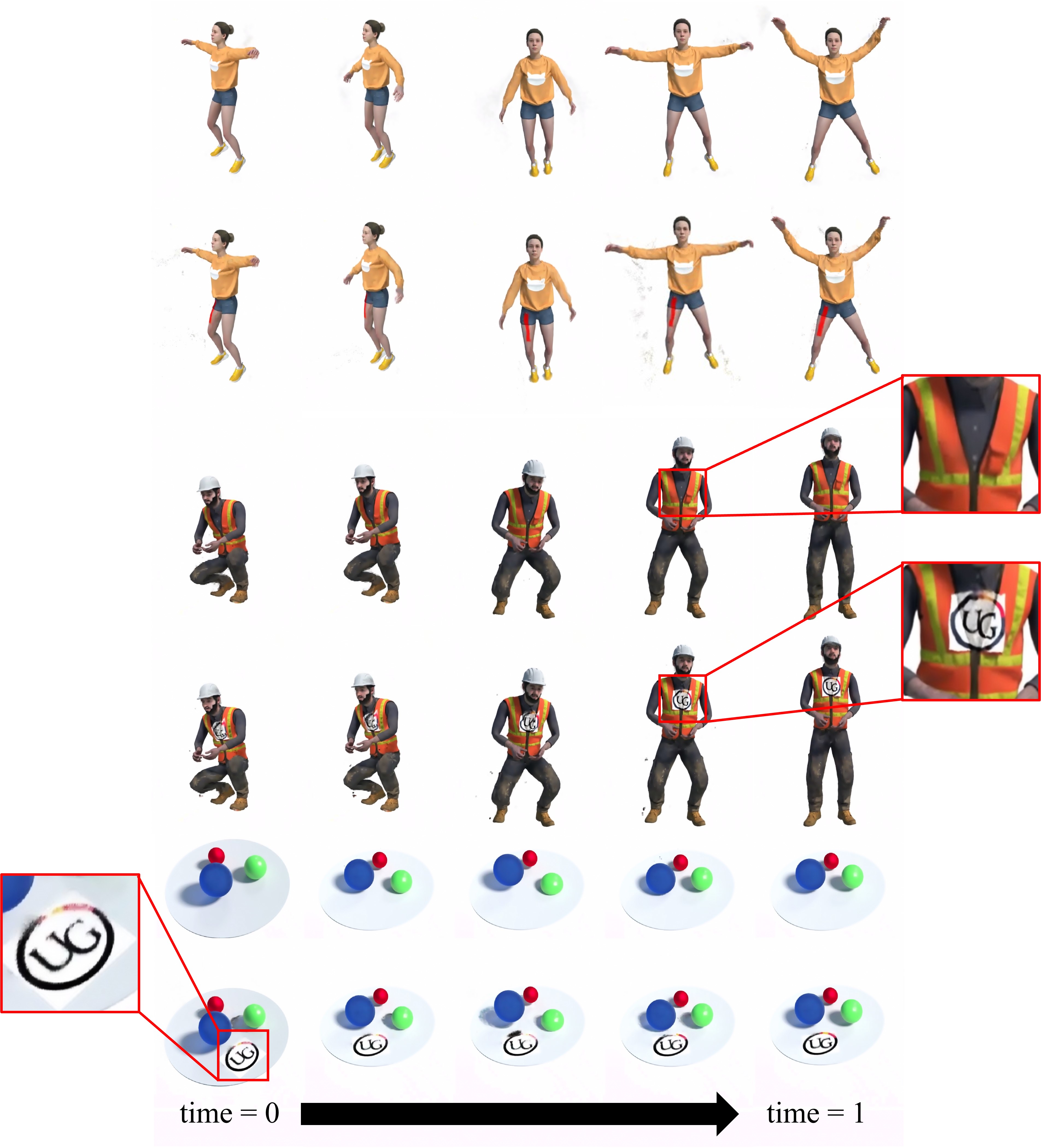}
    \caption{Examples of the brush/sealing tool editing on the scenes: "Jumping Jacks", "Stand Up", and "Bouncing Balls".}
    \label{fig:seald}
\end{figure*}

\section{Experiments}
In this section, the implementation details, evaluation of the implemented D-NeRF, and visualization result of the SealD-NeRF are presented.

\subsection{Implementation Details}
Instant-NGP \cite{muller2022instant} has been selected as the backbone of the framework. The implementation is based on the open-source PyTorch implementation Torch-NGP \cite{Torch-NGP} and Seal-3D \cite{wang2023seal}. All experiments are run on a single Nvidia RTX 4090 GPU. For the D-NeRF training process, the initial learning rate is set to $5e^{-4}$ and exponential decay to $5e^{-5}$ as the same as D-NeRF \cite{pumarola2021d}. Both the canonical network and the deformation network are 8-layers MLPs with ReLU activations. For all the experiments the canonical space is set at $t=0$ by enforcing the deformation to be zero. The training/validation split is the same as the D-NeRF paper. The model is trained with 800 $\times$ 800 images during 300k iterations with a batch size of 4,096 rays. Each ray is sampled 64 times. For the student training process, $\lambda_1 = \lambda_2 = 1$, the same as Seal-3D. The learning rate is set to $5e-4$. Figure \ref{fig:gui} shows an example D-NeRF model and the user interface on the left.

\subsection{D-NeRF}
Table \ref{tab:denerf} summarizes the quantitative results on the 8 dynamic scenes of the dataset released by D-NeRF \cite{pumarola2021d}. The evaluation employs Peak Signal-to-Noise Ratio (PSNR), where higher values indicate better performance. Comparison between the implemented D-NeRF based on the Torch-NGP framework and the original results reported by the D-NeRF paper are presented. The key distinctions between the two implementations are as follows: 1) The original study reduces the input image size to 400 $\times$ 400, whereas our approach maintains the higher resolution of 800 $\times$ 800. 2) The original utilizes CUDA's advanced rendering functions, in contrast to our use of PyTorch's capabilities. 3) The framework provided by the D-NeRF authors includes a ``tv loss'' that aims to minimize the variation in deformation between the current and adjacent time frames, a feature we omitted in our version. Notably, this ``tv loss'' is also absent from the discussion in their published paper.

\begin{table}[]
    \centering
    \caption{The quantitative comparison on the dataset released by D-NeRF.}
    \label{tab:denerf}
    \begin{tabular}{c|c|c}
        Scenes & D-NeRF (Ours) - PSNR & D-NeRF - PSNR \\
        \hline
        Hell Warrior & \textbf{27.33} & 25.02\\
        Mutant & \textbf{36.73} & 31.29\\
        Hook & \textbf{32.36} & 29.25\\
        Bouncing Balls & \textbf{40.08} & 38.93\\
        Lego & \textbf{24.81} & 23.82\\
        T-Rex & 31.59 & \textbf{31.75}\\
        Stand Up & \textbf{37.81} & 32.79\\
        Jumping Jacks & \textbf{33.57} & 32.80\\
    \end{tabular}
\end{table}

Our D-NeRF implementation outperforms the original paper's reported results in all scenes except for the ``T-Rex'' one. This provides strong evidence of the correctness of our implementation. Nonetheless, there are notable differences between the canonical spaces produced by our implementation and those reported in the original paper (shown in Figure \ref{fig:canonical}). This discrepancy does not affect the final outcomes, as editing is not conducted in the canonical space, and the combined operation of the canonical and deformation networks yields accurate results.
\begin{figure*}
    \centering
    \includegraphics[width=1.8\columnwidth]{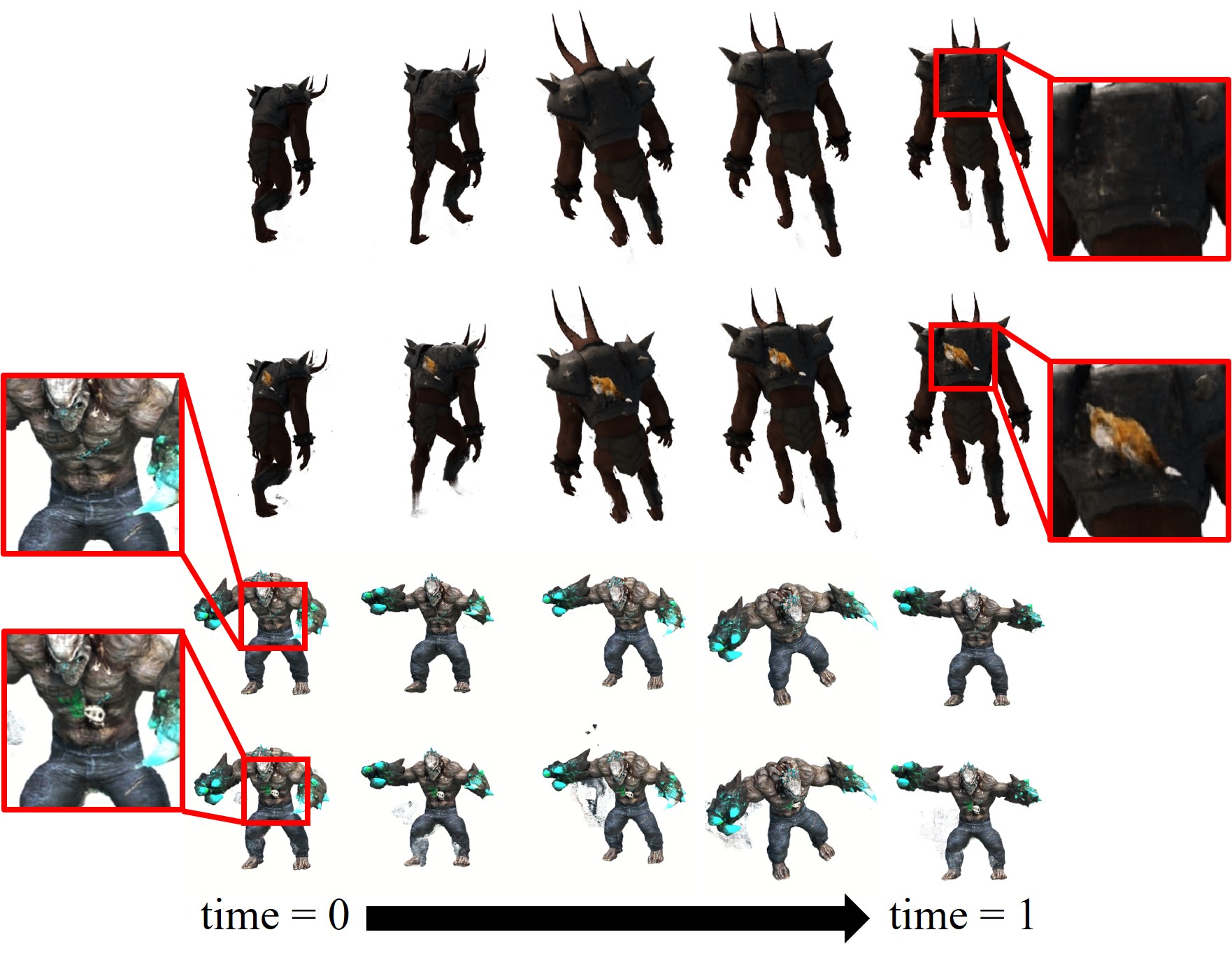}
    \caption{Examples of the sealing tool editing on the scenes: ``Hell Warrior'' and ``Mutant''.}
    \label{fig:trans}
\end{figure*}
\begin{figure}
    \centering
    \includegraphics[width=0.8\columnwidth]{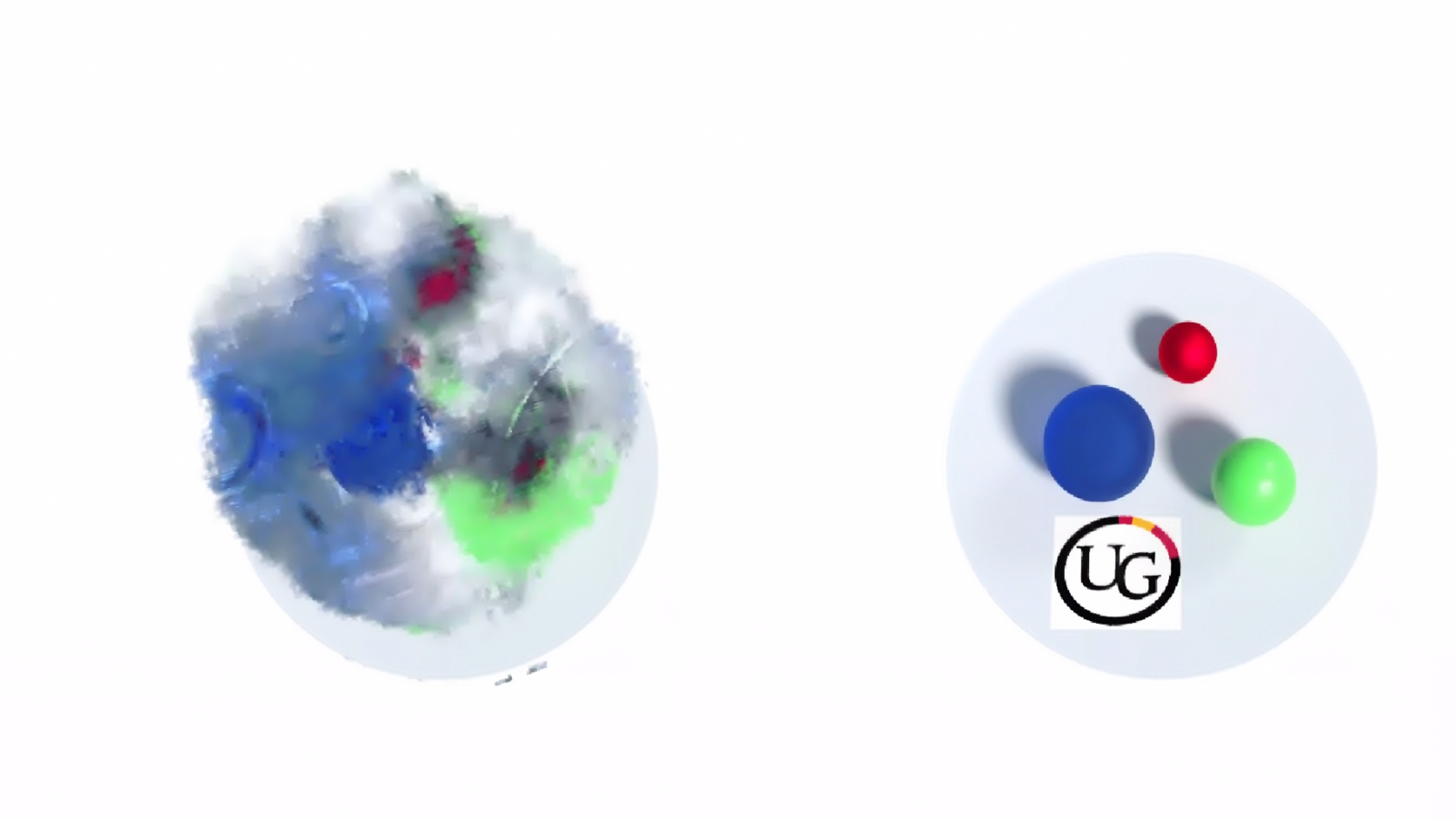}
    \caption{The early stage training result of the student model (left) learning from the teacher model (right).}
    \label{fig:early}
\end{figure}
\subsection{SealD-NeRF}

Figure \ref{fig:seald} illustrates three edits in dynamic scenes. The first involves adding a red brush strip to a leg of a jumping person, which remains attached to the leg throughout the jump. The second edit entails placing a logo image on the chest of a person, which stays fixed to the chest as the person stands up. The third edit involves affixing the logo to a static plate, where the logo remains constant on the plate over time. The first two examples have demonstrated that the proposed method performs well when editing on moving surfaces, while the third example shows that the edit on the static surface acts similarly to Seal-3D which only focuses on static NeRF scenes. Figure \ref{fig:trans} further displays two dynamic scenes edited with images that have transparent backgrounds, demonstrating the capability of our sealing tool to accommodate patterns of any shape. In both examples, the images consistently adhere to the object's surface throughout the entire sequence.

\subsection{Limitations}

The proposed method has a few notable limitations: 1) The student model's learning is based on the teacher model rather than the ground truth, which means any inaccuracies in the teacher model could be transferred. 2) During the initial phase of training, there's a tendency to disrupt the entire scene space, not just the targeted editing area, unlike what is observed with Seal-3D. This suggests that introducing new elements into the canonical space might necessitate relearning, even when the deformation network remains unchanged across all tests. An example of these early-stage training outcomes is illustrated in Figure \ref{fig:early}.

\section{Conclusion and Future Work}
This paper introduces a technique for pixel-level editing in dynamic scenes, building upon the Seal-3D framework within the context of D-NeRF. Currently, two tools are provided: a brush tool and a seal tool, along with a user-friendly interface for editing purposes. Additionally, D-NeRF has been reimplemented to facilitate the proposed method, with both the evaluation of this implementation and the visual outcomes of scene edits being discussed.

Future work could take two primary directions. The first involves expanding the editing toolkit and examining ways to alter the deformation module. A-NeRF \cite{su2021nerf} introduces a technique for pose transfer by estimating the scene's human skeleton, with the main challenge lying in skeletonizing scene objects and linking each volume element to the skeleton. The second direction considers adopting a new foundational model beyond D-NeRF. 3D Gaussian Splatting \cite{kerbl20233d} has recently emerged as a notable advancement in radiance field rendering, offering improvements in render quality and computational efficiency. Its representation of scenes through discrete Gaussians simplifies editing processes, as demonstrated by Gaussianeditor \cite{chen2023gaussianeditor} in static scene editing, making it a promising avenue for future exploration.

\bibliographystyle{IEEEtran}
\bibliography{IEEEtranBST/IEEEabrv,ref}

\begin{thebibliography}{10}
\providecommand{\url}[1]{#1}
\csname url@samestyle\endcsname
\providecommand{\newblock}{\relax}
\providecommand{\bibinfo}[2]{#2}
\providecommand{\BIBentrySTDinterwordspacing}{\spaceskip=0pt\relax}
\providecommand{\BIBentryALTinterwordstretchfactor}{4}
\providecommand{\BIBentryALTinterwordspacing}{\spaceskip=\fontdimen2\font plus
\BIBentryALTinterwordstretchfactor\fontdimen3\font minus \fontdimen4\font\relax}
\providecommand{\BIBforeignlanguage}[2]{{%
\expandafter\ifx\csname l@#1\endcsname\relax
\typeout{** WARNING: IEEEtran.bst: No hyphenation pattern has been}%
\typeout{** loaded for the language `#1'. Using the pattern for}%
\typeout{** the default language instead.}%
\else
\language=\csname l@#1\endcsname
\fi
#2}}
\providecommand{\BIBdecl}{\relax}
\BIBdecl

\bibitem{mildenhall2021nerf}
B.~Mildenhall, P.~P. Srinivasan, M.~Tancik, J.~T. Barron, R.~Ramamoorthi, and R.~Ng, ``Nerf: Representing scenes as neural radiance fields for view synthesis,'' \emph{Communications of the ACM}, vol.~65, no.~1, pp. 99--106, 2021.

\bibitem{pumarola2021d}
A.~Pumarola, E.~Corona, G.~Pons-Moll, and F.~Moreno-Noguer, ``D-nerf: Neural radiance fields for dynamic scenes,'' in \emph{Proceedings of the IEEE/CVF Conference on Computer Vision and Pattern Recognition}, 2021, pp. 10\,318--10\,327.

\bibitem{barron2021mip}
J.~T. Barron, B.~Mildenhall, M.~Tancik, P.~Hedman, R.~Martin-Brualla, and P.~P. Srinivasan, ``Mip-nerf: A multiscale representation for anti-aliasing neural radiance fields,'' in \emph{Proceedings of the IEEE/CVF International Conference on Computer Vision}, 2021, pp. 5855--5864.

\bibitem{zhang2020nerf++}
K.~Zhang, G.~Riegler, N.~Snavely, and V.~Koltun, ``Nerf++: Analyzing and improving neural radiance fields,'' \emph{arXiv preprint arXiv:2010.07492}, 2020.

\bibitem{fridovich2022plenoxels}
S.~Fridovich-Keil, A.~Yu, M.~Tancik, Q.~Chen, B.~Recht, and A.~Kanazawa, ``Plenoxels: Radiance fields without neural networks,'' in \emph{Proceedings of the IEEE/CVF Conference on Computer Vision and Pattern Recognition}, 2022, pp. 5501--5510.

\bibitem{muller2022instant}
T.~M{\"u}ller, A.~Evans, C.~Schied, and A.~Keller, ``Instant neural graphics primitives with a multiresolution hash encoding,'' \emph{ACM Transactions on Graphics (ToG)}, vol.~41, no.~4, pp. 1--15, 2022.

\bibitem{cao2023hexplane}
A.~Cao and J.~Johnson, ``Hexplane: A fast representation for dynamic scenes,'' in \emph{Proceedings of the IEEE/CVF Conference on Computer Vision and Pattern Recognition}, 2023, pp. 130--141.

\bibitem{liu2022devrf}
J.-W. Liu, Y.-P. Cao, W.~Mao, W.~Zhang, D.~J. Zhang, J.~Keppo, Y.~Shan, X.~Qie, and M.~Z. Shou, ``Devrf: Fast deformable voxel radiance fields for dynamic scenes,'' \emph{Advances in Neural Information Processing Systems}, vol.~35, pp. 36\,762--36\,775, 2022.

\bibitem{xian2021space}
W.~Xian, J.-B. Huang, J.~Kopf, and C.~Kim, ``Space-time neural irradiance fields for free-viewpoint video,'' in \emph{Proceedings of the IEEE/CVF Conference on Computer Vision and Pattern Recognition}, 2021, pp. 9421--9431.

\bibitem{park2021nerfies}
K.~Park, U.~Sinha, J.~T. Barron, S.~Bouaziz, D.~B. Goldman, S.~M. Seitz, and R.~Martin-Brualla, ``Nerfies: Deformable neural radiance fields,'' in \emph{Proceedings of the IEEE/CVF International Conference on Computer Vision}, 2021, pp. 5865--5874.

\bibitem{liu2021editing}
S.~Liu, X.~Zhang, Z.~Zhang, R.~Zhang, J.-Y. Zhu, and B.~Russell, ``Editing conditional radiance fields,'' in \emph{Proceedings of the IEEE/CVF international conference on computer vision}, 2021, pp. 5773--5783.

\bibitem{yang2021learning}
B.~Yang, Y.~Zhang, Y.~Xu, Y.~Li, H.~Zhou, H.~Bao, G.~Zhang, and Z.~Cui, ``Learning object-compositional neural radiance field for editable scene rendering,'' in \emph{Proceedings of the IEEE/CVF International Conference on Computer Vision}, 2021, pp. 13\,779--13\,788.

\bibitem{kuang2023palettenerf}
Z.~Kuang, F.~Luan, S.~Bi, Z.~Shu, G.~Wetzstein, and K.~Sunkavalli, ``Palettenerf: Palette-based appearance editing of neural radiance fields,'' in \emph{Proceedings of the IEEE/CVF Conference on Computer Vision and Pattern Recognition}, 2023, pp. 20\,691--20\,700.

\bibitem{liu2022nerf}
H.-K. Liu, I.~Shen, B.-Y. Chen \emph{et~al.}, ``Nerf-in: Free-form nerf inpainting with rgb-d priors,'' \emph{arXiv preprint arXiv:2206.04901}, 2022.

\bibitem{wang2023seal}
X.~Wang, J.~Zhu, Q.~Ye, Y.~Huo, Y.~Ran, Z.~Zhong, and J.~Chen, ``Seal-3d: Interactive pixel-level editing for neural radiance fields,'' in \emph{Proceedings of the IEEE/CVF International Conference on Computer Vision}, 2023, pp. 17\,683--17\,693.

\bibitem{weng2022humannerf}
C.-Y. Weng, B.~Curless, P.~P. Srinivasan, J.~T. Barron, and I.~Kemelmacher-Shlizerman, ``Humannerf: Free-viewpoint rendering of moving people from monocular video,'' in \emph{Proceedings of the IEEE/CVF conference on computer vision and pattern Recognition}, 2022, pp. 16\,210--16\,220.

\bibitem{fang2022fast}
J.~Fang, T.~Yi, X.~Wang, L.~Xie, X.~Zhang, W.~Liu, M.~Nie{\ss}ner, and Q.~Tian, ``Fast dynamic radiance fields with time-aware neural voxels,'' in \emph{SIGGRAPH Asia 2022 Conference Papers}, 2022, pp. 1--9.

\bibitem{gao2021dynamic}
C.~Gao, A.~Saraf, J.~Kopf, and J.-B. Huang, ``Dynamic view synthesis from dynamic monocular video,'' in \emph{Proceedings of the IEEE/CVF International Conference on Computer Vision}, 2021, pp. 5712--5721.

\bibitem{li2021neural}
Z.~Li, S.~Niklaus, N.~Snavely, and O.~Wang, ``Neural scene flow fields for space-time view synthesis of dynamic scenes,'' in \emph{Proceedings of the IEEE/CVF Conference on Computer Vision and Pattern Recognition}, 2021, pp. 6498--6508.

\bibitem{zeng2023efficient}
J.~Zeng, Y.~Li, Y.~Ran, S.~Li, F.~Gao, L.~Li, S.~He, J.~Chen, and Q.~Ye, ``Efficient view path planning for autonomous implicit reconstruction,'' in \emph{2023 IEEE International Conference on Robotics and Automation (ICRA)}.\hskip 1em plus 0.5em minus 0.4em\relax IEEE, 2023, pp. 4063--4069.

\bibitem{li2020inverse}
Z.~Li, M.~Shafiei, R.~Ramamoorthi, K.~Sunkavalli, and M.~Chandraker, ``Inverse rendering for complex indoor scenes: Shape, spatially-varying lighting and svbrdf from a single image,'' in \emph{Proceedings of the IEEE/CVF Conference on Computer Vision and Pattern Recognition}, 2020, pp. 2475--2484.

\bibitem{park2021hypernerf}
K.~Park, U.~Sinha, P.~Hedman, J.~T. Barron, S.~Bouaziz, D.~B. Goldman, R.~Martin-Brualla, and S.~M. Seitz, ``Hypernerf: A higher-dimensional representation for topologically varying neural radiance fields,'' \emph{arXiv preprint arXiv:2106.13228}, 2021.

\bibitem{guo2020object}
M.~Guo, A.~Fathi, J.~Wu, and T.~Funkhouser, ``Object-centric neural scene rendering,'' \emph{arXiv preprint arXiv:2012.08503}, 2020.

\bibitem{yu2022monosdf}
Z.~Yu, S.~Peng, M.~Niemeyer, T.~Sattler, and A.~Geiger, ``Monosdf: Exploring monocular geometric cues for neural implicit surface reconstruction,'' \emph{Advances in neural information processing systems}, vol.~35, pp. 25\,018--25\,032, 2022.

\bibitem{hasselgren2022shape}
J.~Hasselgren, N.~Hofmann, and J.~Munkberg, ``Shape, light, and material decomposition from images using monte carlo rendering and denoising,'' \emph{Advances in Neural Information Processing Systems}, vol.~35, pp. 22\,856--22\,869, 2022.

\bibitem{wang2021neus}
P.~Wang, L.~Liu, Y.~Liu, C.~Theobalt, T.~Komura, and W.~Wang, ``Neus: Learning neural implicit surfaces by volume rendering for multi-view reconstruction,'' \emph{arXiv preprint arXiv:2106.10689}, 2021.

\bibitem{yang2022neumesh}
B.~Yang, C.~Bao, J.~Zeng, H.~Bao, Y.~Zhang, Z.~Cui, and G.~Zhang, ``Neumesh: Learning disentangled neural mesh-based implicit field for geometry and texture editing,'' in \emph{European Conference on Computer Vision}.\hskip 1em plus 0.5em minus 0.4em\relax Springer, 2022, pp. 597--614.

\bibitem{Torch-NGP}
``Torch-ngp github repository,'' \url{https://github.com/ashawkey/torch-ngp}.

\bibitem{su2021nerf}
S.-Y. Su, F.~Yu, M.~Zollh{\"o}fer, and H.~Rhodin, ``A-nerf: Articulated neural radiance fields for learning human shape, appearance, and pose,'' \emph{Advances in Neural Information Processing Systems}, vol.~34, pp. 12\,278--12\,291, 2021.

\bibitem{kerbl20233d}
B.~Kerbl, G.~Kopanas, T.~Leimk{\"u}hler, and G.~Drettakis, ``3d gaussian splatting for real-time radiance field rendering,'' \emph{ACM Transactions on Graphics}, vol.~42, no.~4, 2023.

\bibitem{chen2023gaussianeditor}
Y.~Chen, Z.~Chen, C.~Zhang, F.~Wang, X.~Yang, Y.~Wang, Z.~Cai, L.~Yang, H.~Liu, and G.~Lin, ``Gaussianeditor: Swift and controllable 3d editing with gaussian splatting,'' \emph{arXiv preprint arXiv:2311.14521}, 2023.

\end{thebibliography}



% \end{thebibliography}

\end{document}